\documentclass{article}

\usepackage[final]{nips_2017}

\usepackage[utf8]{inputenc} 
\usepackage[T1]{fontenc}    
\usepackage{hyperref}       
\usepackage{url}            
\usepackage{booktabs}       
\usepackage{amsfonts}       
\usepackage{nicefrac}       
\usepackage{microtype}      
\usepackage{graphicx} 
\usepackage{caption}
\usepackage{subcaption}
\usepackage{amsmath,amsthm,amssymb}
\usepackage{natbib}
\usepackage{enumitem,kantlipsum}
\usepackage{placeins}
\usepackage{pdfpages}

\newcommand{\argmin}{\operatornamewithlimits{argmin}}

\setcitestyle{numbers,square}

\title{Robustly representing uncertainty through sampling\\ in deep neural networks}

\author{
  Patrick McClure \\
  MRC Cognition and Brain Sciences Unit\\
  University of Cambridge\\
  \texttt{patrick.mcclure@mrc-cbu.cam.ac.uk} \\
  \And
  Nikolaus Kriegeskorte \\
  Department of Psychology\\
  Columbia University\\
  \texttt{nk2765@columbia.edu} \\
}

\begin{document}

\maketitle

\begin{center} {\bf Abstract} \end{center}
As deep neural networks (DNNs) are applied to increasingly challenging problems, they will need to be able to represent their own uncertainty. Modeling uncertainty is one of the key features of Bayesian methods. Using Bernoulli dropout with sampling at prediction time has recently been proposed as an efficient and well performing variational inference method for DNNs. However, sampling from other multiplicative noise based variational distributions has not been investigated in depth. We evaluated Bayesian DNNs trained with Bernoulli or Gaussian multiplicative masking of either the units (dropout) or the weights (dropconnect). We tested the calibration of the probabilistic predictions of Bayesian convolutional neural networks (CNNs) on MNIST and CIFAR-10. Sampling at prediction time increased the calibration of the DNNs' probabalistic predictions. Sampling weights, whether Gaussian or Bernoulli, led to more robust representation of uncertainty compared to sampling of units. However, using either Gaussian or Bernoulli dropout led to increased test set classification accuracy. Based on these findings we used both Bernoulli dropout and Gaussian dropconnect concurrently, which we show approximates the use of a spike-and-slab variational distribution without increasing the number of learned parameters. We found that spike-and-slab sampling had higher test set performance than Gaussian dropconnect and more robustly represented its uncertainty compared to Bernoulli dropout.

\section{Introduction}
\label{introduction}

Deep neural networks (DNNs), particularly convolutional neural networks (CNNs), have recently been used to solve complex perceptual and decision tasks \cite{Krizhevsky2012,Mnih2015,silver2016mastering}. While these models take into account aleatoric uncertainty via their softmax output (i.e. the uncertainty present in the training data), they do not take into account epistemic uncertainty (i.e. parameter uncertainty) \cite{kendall2017uncertainties}. Bayesian DNNs attempt to learn a distribution over their parameters thereby allowing for the computation of the uncertainty of their outputs given the parameters. However, ideal Bayesian methods do not scale well due to the difficulty in computing the posterior of a network's parameters. 

As a result, several approximate Bayesian methods have been proposed for DNNs. Using the Laplace approximation was proposed by \cite{mackay1992practical}. Using Markov chain Monte Carlo (MCMC) has been suggested to estimate the posterior of the networks weights given the training data \cite{neal2012bayesian,welling2011bayesian} . Using expectation propagation has also been proposed \cite{jylanki2014expectation, hernandez2015probabilistic}. However, these methods can be difficult to implement for the very large CNNs commonly used for object recognition. Variational inference methods have also been used to make Bayesian NNs more tractable\cite{hinton1993keeping,barber1998ensemble,graves2011practical,blundell2015weight}. Due in large part to the fact that these methods substantially increase the number of parameters in a network, they have not been extensively applied to large DNNs. \citet{gal2016bayesian} and \citet{kingma2015variational} bypassed this issue by developing Bayesian CNNs using Bernoulli and Gaussian dropout \cite{srivastava2014dropout}, respectively. While independent weight sampling with additive Gaussian noise has been investigated \cite{hinton1993keeping,barber1998ensemble,graves2011practical,blundell2015weight}, independently sampling weights using multiplicative Bernoulli noise, i.e. dropconnect \cite{wan2013regularization}, or independently sampled multiplicative Gaussian noise has not been thoroughly evaluated.

In addition to Bernoulli and Gaussian distributions, spike-and-slab distributions, a combination of the two, have been investigated, particularly for linear models \cite{mitchell1988bayesian,madigan1994model,george1997approaches,ishwaran2005spike}. Interestingly, Bernoulli dropout and dropconnect can be seen as approximations to spike-and-slab distributions for units and weights, respectively \cite{louizos2015smart,Gal2016Uncertainty}. Spike-and-slab variational distributions have been implemented using Bernoulli dropout with additive weight noise sampled from a Gaussian with a learned standard deviation \cite{louizos2015smart}. This approach more than doubled the number of learned parameters, since the mean and the standard deviation of each weight as well as the dropout rate for each unit were learned. However, this method did not consistently outperform standard neural networks. \citet{Gal2016Uncertainty} also discussed motivations for spike-and-slab variational distributions, but did not suggest a practical implementation.

We evaluated the performance Bayesian CNNs with different variational distributions on MNIST \citep{lecun1998gradient} and CIFAR-10 \cite{krizhevsky2009learning}. We also investigate how adding Gaussian image noise with varying standard deviations to the test set affected each network's learned uncertainty. We did this to test how networks responded to inputs not drawn from the data distribution used to create the training and test sets. We also propose an approximation of the spike-and-slab variational inference based on Bernoulli dropout and Gaussian dropconnect, which combines the advantages of Gaussian dropconnect and Bernoulli dropout sampling leading to better uncertainty estimates and good test set generalization without increasing the number of learned parameters.

\section{Methods}
\label{methods}

\subsection{Bayesian Deep Neural Networks}

DNNs are commonly trained by finding the maximum a posteriori (MAP) weights given the training data ($D_{train}$) and a prior over the weight matrix $W$, $p(W)$. However, ideal Bayesian learning would involve computing the full posterior. This can be intractable due to both the difficulty in calculating $p(D_{train})$ and in calculating the joint distribution of a large number of parameters. Instead, $p(W|D_{train})$ can be approximated using a variational distribution $q(W)$. This distribution is constructed to allow for easy generation of samples. The objective of variational inference is to optimize the variational parameters $V$ so that the Kullblack-Leiber (KL) divergence between $q_V(W)$ and $p(W|D_{train})$ is minimized \cite{hinton1993keeping,barber1998ensemble,graves2011practical,blundell2015weight}:

\begin{equation}
\begin{aligned}
V^* &  = \argmin_{V} KL[q_V(W)||p(W)] - \int q_V(W) \log p(D_{train}|W) dW \\
\end{aligned}
\end{equation}

Using Monte Carlo (MC) methods to estimate $E_{q_V(w)}[\log p(D_{train}|W)]$, using weight samples $ \hat{W}^i \sim q_V(W)$, results in the following loss function:

\begin{equation}
\mathcal{L} := KL(q_V(W)||p(W)) -\frac{1}{n} \sum_i^n \log p(D_{train}|\hat{W}^i)
\end{equation}

MC sampling can also be used to estimate the probability of test data:

\begin{equation}
p(D_{test}) \approx \frac{1}{n} \sum_i^n p(D_{test}|\hat{W}^i)
\end{equation}

\subsection{Variational Distributions}
The number and continuous nature of the parameters in DNNs makes sampling from the entire distribution of possible weight matrices computationally challenging. However, variational distributions can make sampling easier. In deep learning, the most common sampling method is using multiplicative noise masks drawn from some distribution. Several of these methods can be formulated as variational distributions where weights are sampled by element-wise multiplication of the variational parameters $V$, the $n \times n$ connection matrix with an element for each connection between the $n$ units in the network, by a mask $\hat{M}$, which is sampled from some probability distribution:

\begin{equation}
\hat{W} = V \circ \hat{M} \ \ where \ \ \hat{M} \sim p(M) 
\end{equation}  

\begin{figure}
\centering
    \includegraphics[width=0.6\textwidth]{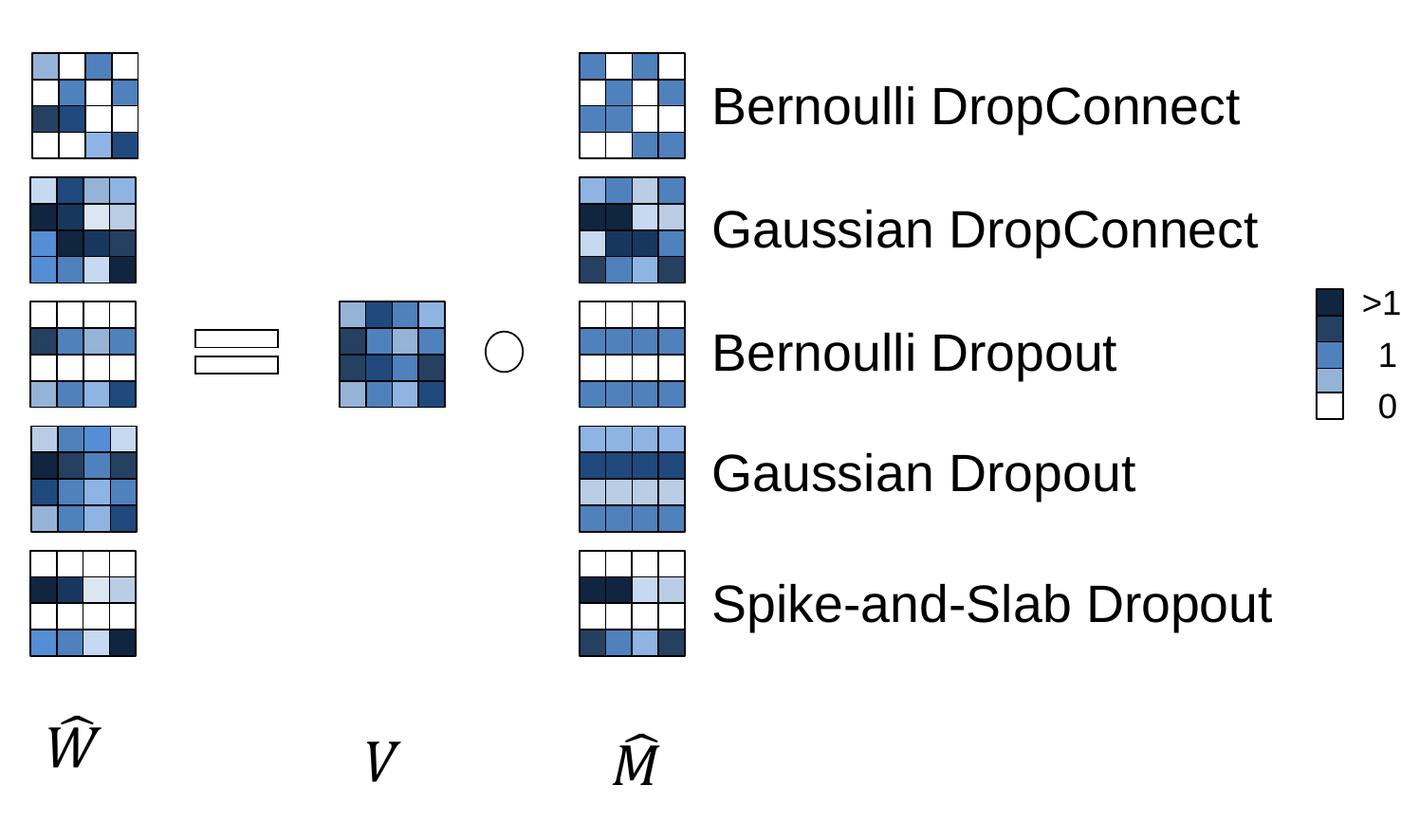}
    \caption{An illustration of sampling network weights using the different variational distributions.}
    \label{sampling_illustration}
\end{figure}

From this perspective, the difference between dropout and dropconnect, as well as Bernoulli and Gaussian methods, is simply the probability distribution used to generate the mask sample (Figure \ref{sampling_illustration}). 

\subsubsection{Bernoulli Dropconnect \& Dropout}
In Bernoulli dropconnect, each element of the mask is sampled independently, so $\hat{m}_{i,j} \sim Bernoulli(1-p)$ where $p$ is the probability of dropping a connection. In Bernoulli dropout, however, the weights are not sampled independently. Instead, one Bernoulli variable is sampled for each row of the weight matrix, so $\hat{m}_{i,*} \sim Bernoulli(1-p)$ where $p$ is the probability of dropping a unit. 

\subsubsection{Gaussian Dropconnect \& Dropout}
In Gaussian dropconnect and dropout, $\hat{w}_{i,j}$ is sampled from a Gaussian distribution centered at variational parameter $v_{i,j}$. This is accomplished by sampling the multiplicative mask using Gaussian distributions with a mean of 1 and a variance of $\sigma_{dc}^2 = p/(1-p)$, which matches the mean and variance of Bernoulli dropout when training time scaling is used \cite{srivastava2014dropout}. In Gaussian dropconnect, each element of the mask is sampled independently, which results in  $\hat{m}_{i,j} \sim\mathcal{N}(1,\sigma_{dc}^2)$. In Gaussian dropout, each element in a row has the same random variable, so $ \hat{m}_{i,*} \sim \mathcal{N}(1,\sigma_{dc}^2)$. It can be shown that using Gaussian dropconnect or dropout with L2-regularization leads to optimizing a stochastic lower-bound of the variational objective function (See Supplementary Material). 

\subsubsection{Spike-and-Slab Dropout}
A spike-and-slab distribution is the normalized linear combination of a "spike" of probability mass at zero and a "slab" consisting of a Gaussian distribution.  This spike-and-slab returns a 0 with probability $p_{spike}$ or a random sample from a Gaussian distribution $\mathcal{N}(\mu_{slab},\sigma_{slab}^2)$ with probability $1 - p_{spike}$. We propose concurrently using Bernoulli dropout and Gaussian dropconnect to approximate the use of a spike-and-slab variational distribution and spike-and-slab prior by optimizing a lower-bound of the variational objective function (See Supplementary Material). In this formulation, $m_{i,j} \sim b_{i,*} \mathcal{N}(1,\sigma_{dc}^2)$, where $ b_{i,*} \sim Bern(1 - p_{do})$ for each mask row and $\sigma_{dc}^2 = p_{dc}/(1-p_{dc})$. As for Bernoulli dropout, each row of the mask $M$ is multiplied by 0 with probability $p_{do}$, otherwise each element in that row is multiplied by a value independently sampled from a Gaussian distribution as in Gaussian dropconnect. During non-sampling inference, spike-and-slab dropout uses the mean weight values and, per Bernoulli dropout, multiplies unit outputs by $1-p_{do}$.

\begin{figure}[t]
    \centering
      \includegraphics[width=\textwidth]{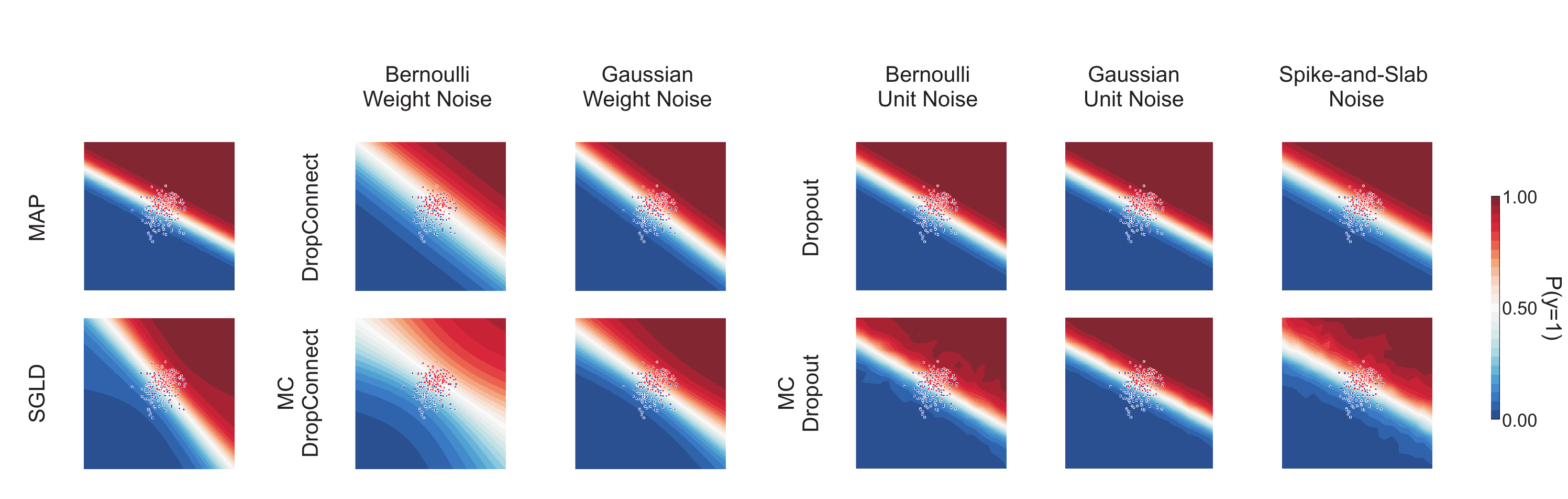}
        \caption{ The probabalistic logistic regression decision boundaries of a linear network for Bernoulli dropconnect (BDC), Gaussian dropconnect (GDC), Bernoulli dropout (BDO), Gaussian dropout (GDO), and spike-and-slab dropout (SSD) with and without MC sampling compared to the decision bondaries for the MAP solution and stochactic gradient Langevin dynamics (SGLD) \cite{welling2011bayesian}.}
        \label{log_reg}
\end{figure}

\section{Experiments}
\label{experiments}

\subsection{Logistic Regression}
In order to visualize the effects of each variational distribution, we trained linear networks with five hidden units to classify data drawn from two 2D multivariate Gaussian distributions. Multiple linear units were used so that Bernoulli dropout would not dropout the only unit in the network. For the dropout methods, unit sampling was performed on the linear hidden layer. For the dropconnect methods, every weight was sampled. Dropout and dropconnect probabilities of $p = 0.4$ were used for each of these networks, except for the spike-and-slab dropconnect probability which was $0.2$. In Figure \ref{log_reg}, we show the decision boundaries learned by the various networks. Higher variability in the decision boundaries corresponds to higher uncertainty. All of the MC sampling methods predict with higher uncertainty as points become further away from the training data. This is particularly true for the dropconnect and spike-and-slab methods.

\begin{table}[b]
\centering
\caption{MNIST and CIFAR-10 mean and standard deviation of test errors for the trained convolutional neural networks (CNNs) with and without Monte-Carlo (MC) across 5 runs, each MC run using 10 samples.}
\label{cnn-test-errors}
\vskip 0.15in
\begin{scriptsize}
\begin{tabular}{c c c c c}
 & \multicolumn{2}{c}{\textbf{MNIST}}  & \multicolumn{2}{c}{\textbf{CIFAR-10}}  \\
\hline
\textbf{Method} & \textbf{ Mean Error (\%)} & \textbf{Error Std. Dev.} & \textbf{ Mean Error (\%)} & \textbf{Error Std. Dev.} \\
\hline
MAP    & 0.76 & \-- & 25.86 & \-- \\
Bernoulli DropConnect    & 0.56 & \-- & 16.46 & \-- \\
MC Bernoulli DropConnect & 0.56 & 0.03 & 16.59 & 0.11 \\
Gaussian DropConnect & 0.56 & \-- & 16.78 & \-- \\
MC Gaussian DropConnect    & 0.58 & 0.02 &  16.65 & 0.11 \\
Bernoulli Dropout    & 0.49 & \-- & 11.23 & \-- \\
MC Bernoulli Dropout & 0.48 & 0.03 & 9.95 & 0.08 \\
Gaussian Dropout & 0.42 & \-- & 9.07 & \-- \\
MC Gaussian Dropout    & 0.36 & 0.04 & 9.00 & 0.10 \\
Spike-and-Slab Dropout & 0.48  & -- & 10.64  & -- \\
MC Spike-and-Slab Dropout & 0.46 & 0.01 & 10.05 & 0.06 \\
\hline
\end{tabular}
\end{scriptsize}
\vskip -0.1in
\end{table}

\subsection{Convolutional Neural Networks}
We trained CNNs on MNIST \citep{lecun1998gradient} and CIFAR-10 \cite{krizhevsky2009learning}. For each dataset, a 10,000 image subset of the training set was used for validation. For MNIST, each CNN had two convolutional layers followed by a fully connected layer and a softmax layer. For CIFAR-10, each CNN had 13 convolutional layers followed by a fully connected layer and a softmax layer. (See Supplementary Material for the detailed architectures.) For the dropout networks, dropout was used after each convolutional and fully-connected layer, but before the non-linearity. For the dropconnect networks, all weights were sampled. All $p$s were treated as network-wide hyperparameters. For L2-regularization, L2-coefficients of 1e-5 (MNIST) and 4e-5 (CIFAR-10) were used for all weights. No data augmentation was used for MNIST. Random horizontal flipping was used during CIFAR-10 training. We evaluated the trained CNNs using the original testing sets and using the testing images with added random Gaussian noise of increasing variance in order to test each network's uncertainty for the regions of input space not seen in the training set.

\begin{figure}
    \centering
    \begin{subfigure}[b]{0.3\textwidth}
        \includegraphics[width=\textwidth]{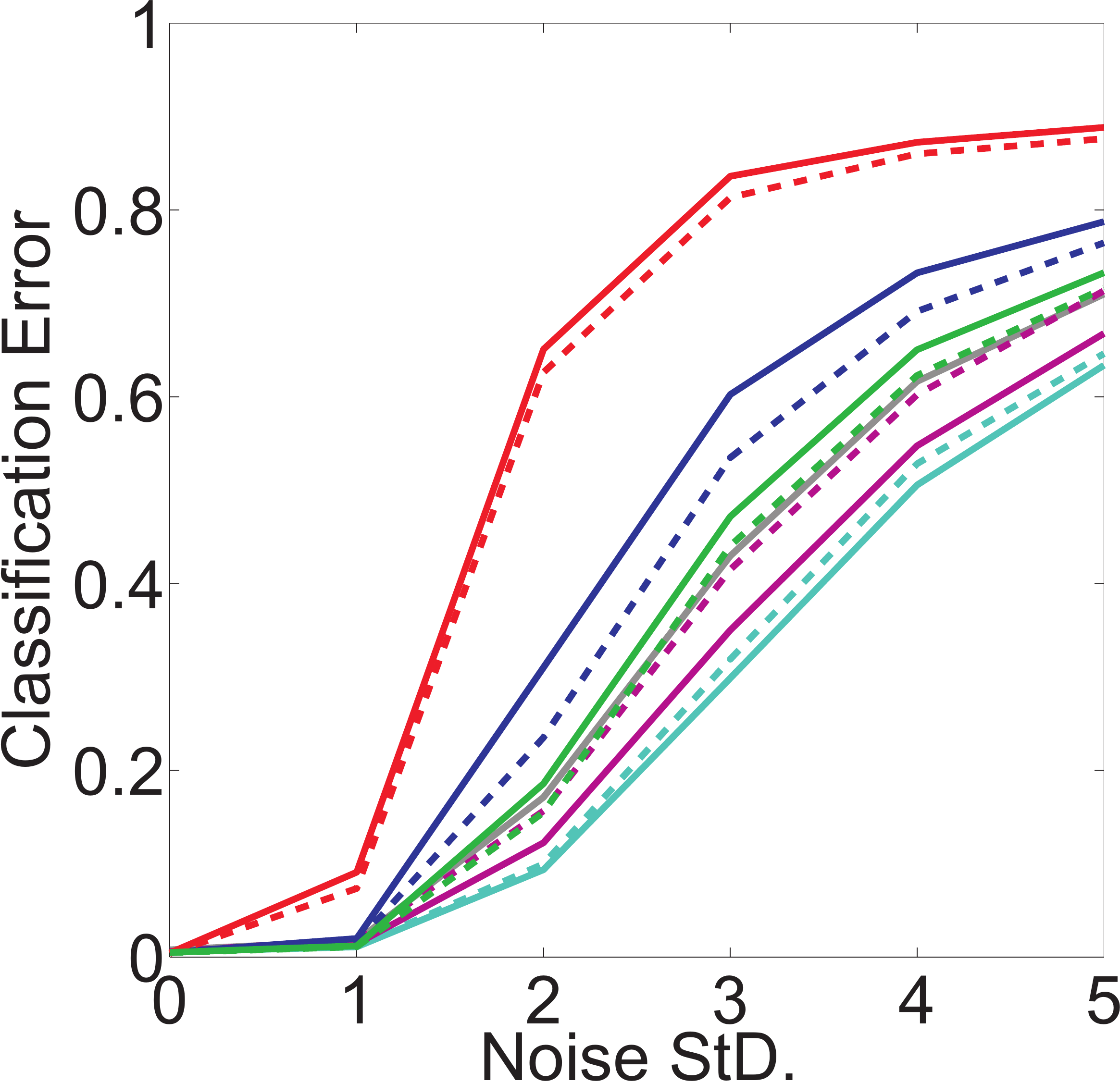}
        \caption{}
        \label{mnist_error_vs_noise}
    \end{subfigure}
    ~ 
    \begin{subfigure}[b]{0.3\textwidth}
        \includegraphics[width=\textwidth]{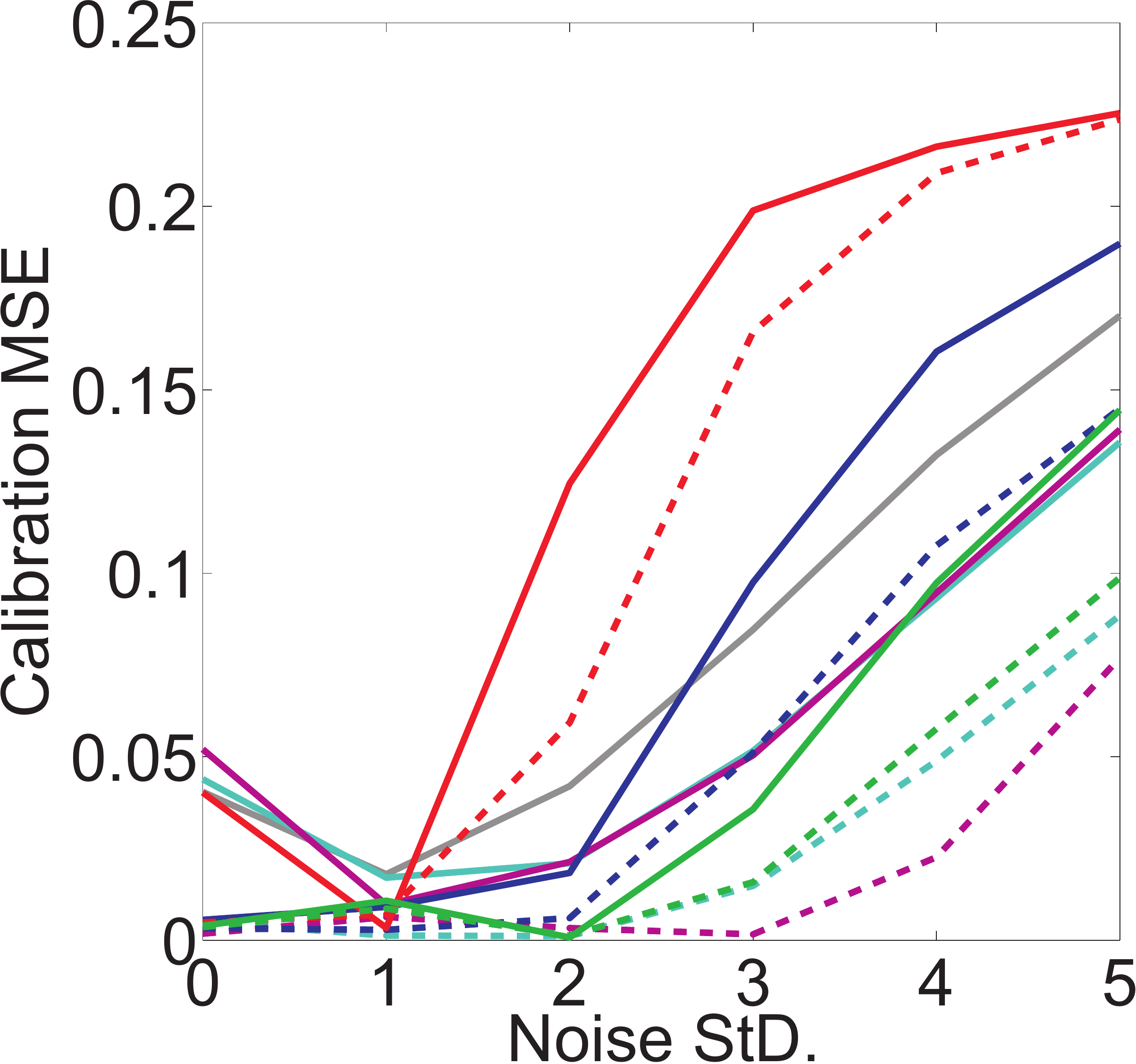}
        \caption{}
        \label{mnist_calibration_vs_noise}
    \end{subfigure}
    ~ 
    \begin{subfigure}[b]{0.3\textwidth}
        \includegraphics[width=\textwidth]{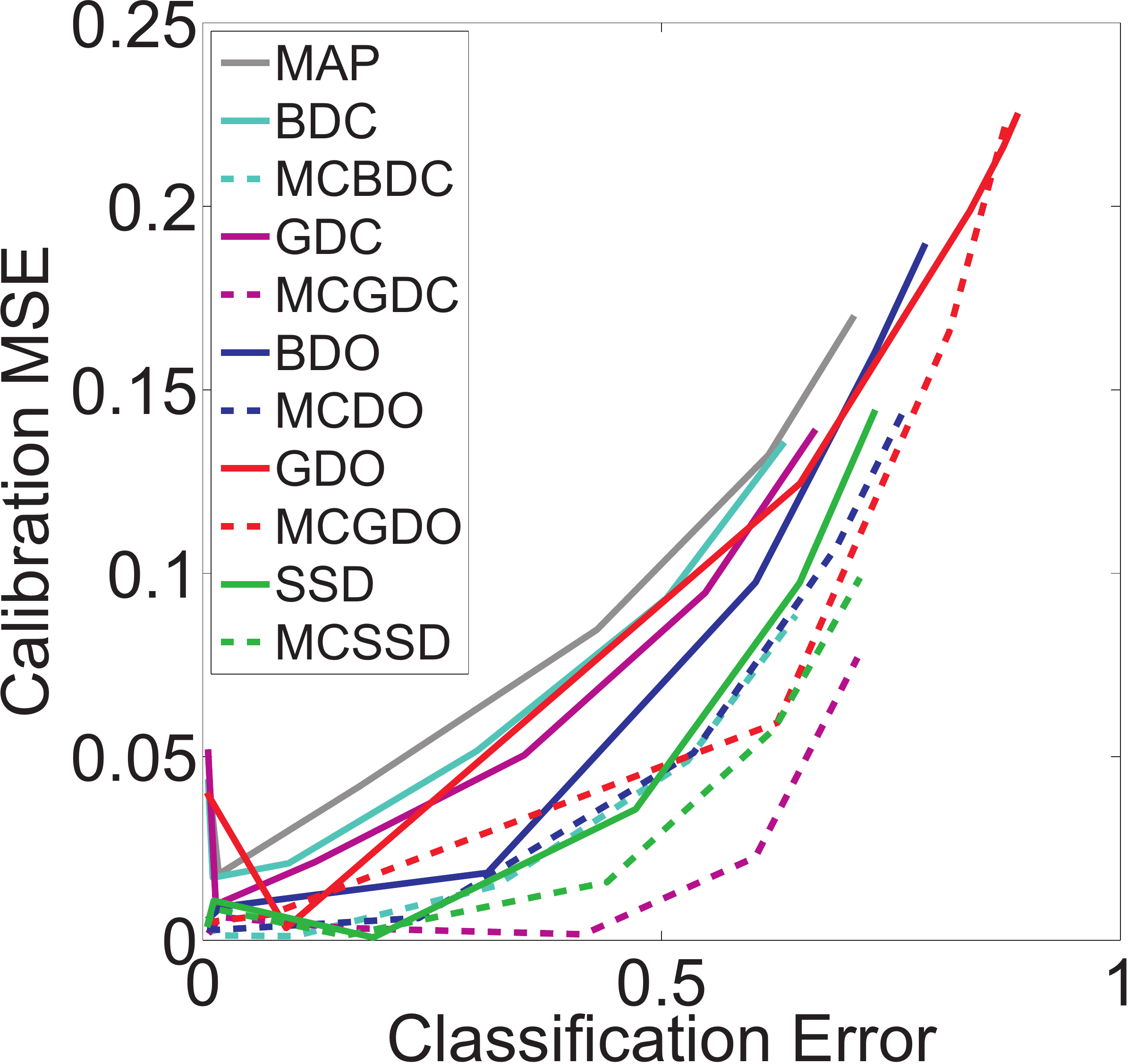}
        \caption{}
        \label{mnist_calibration_vs_error}
    \end{subfigure}
    \caption{The MNIST (a) classification error for additive Gaussian noise using standard deviations St.D of 0, 1, 2, 3, 4, and 5, (b) mean squared error (MSE) between the $x = y$ line and the calibration plot (i.e. the frequency of the true label vs predicted probability of that label) for varying Gaussian image noise StD., and (c) calibration MSE versus the classification error for predicitons across all noise StD. for Bernoulli dropconnect (BDC), Gaussian dropconnect (GDC), Bernoulli dropout (BDO), Gaussian dropout (GDO), and spike-and-slab dropout (SSD) with and without MC sampling using 10 samples.}\label{mnist_plots}
\end{figure}

\begin{figure}
    \centering
    \begin{subfigure}[b]{0.3\textwidth}
        \includegraphics[width=\textwidth]{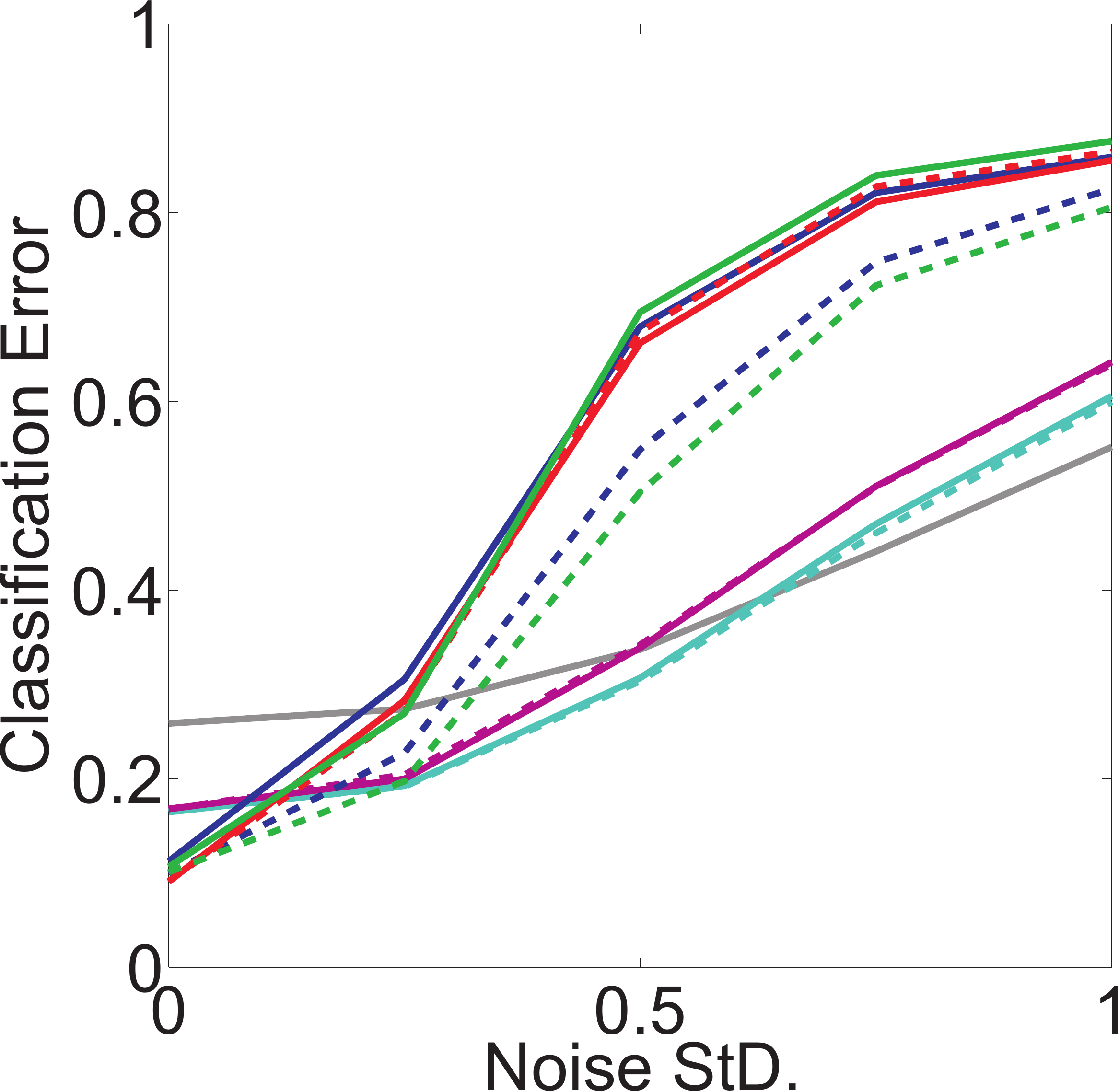}
        \caption{}
        \label{error_vs_noise}
    \end{subfigure}
    ~ 
    \begin{subfigure}[b]{0.3\textwidth}
        \includegraphics[width=\textwidth]{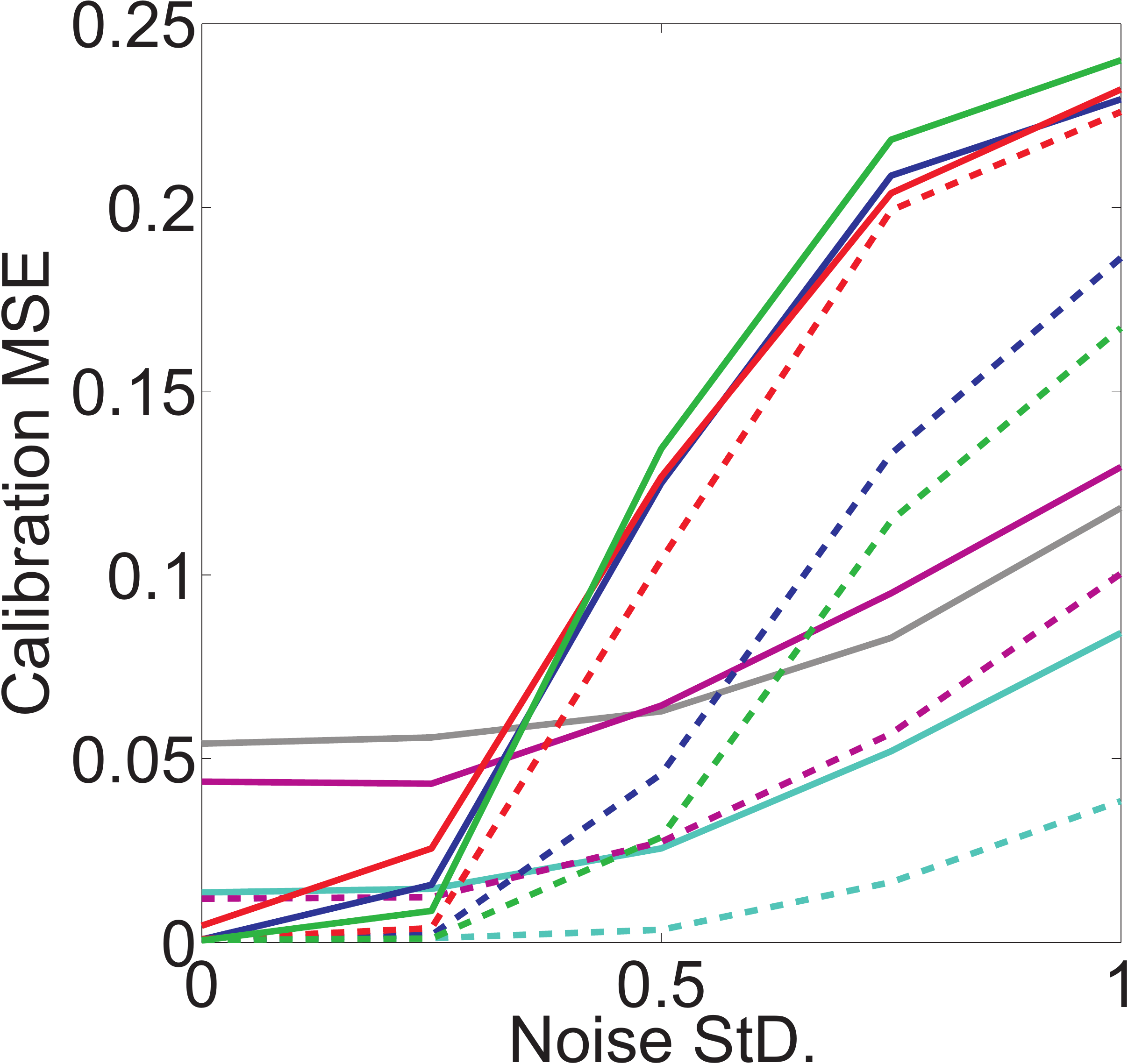}
        \caption{}
        \label{calibration_vs_noise}
    \end{subfigure}
    ~ 
    \begin{subfigure}[b]{0.3\textwidth}
        \includegraphics[width=\textwidth]{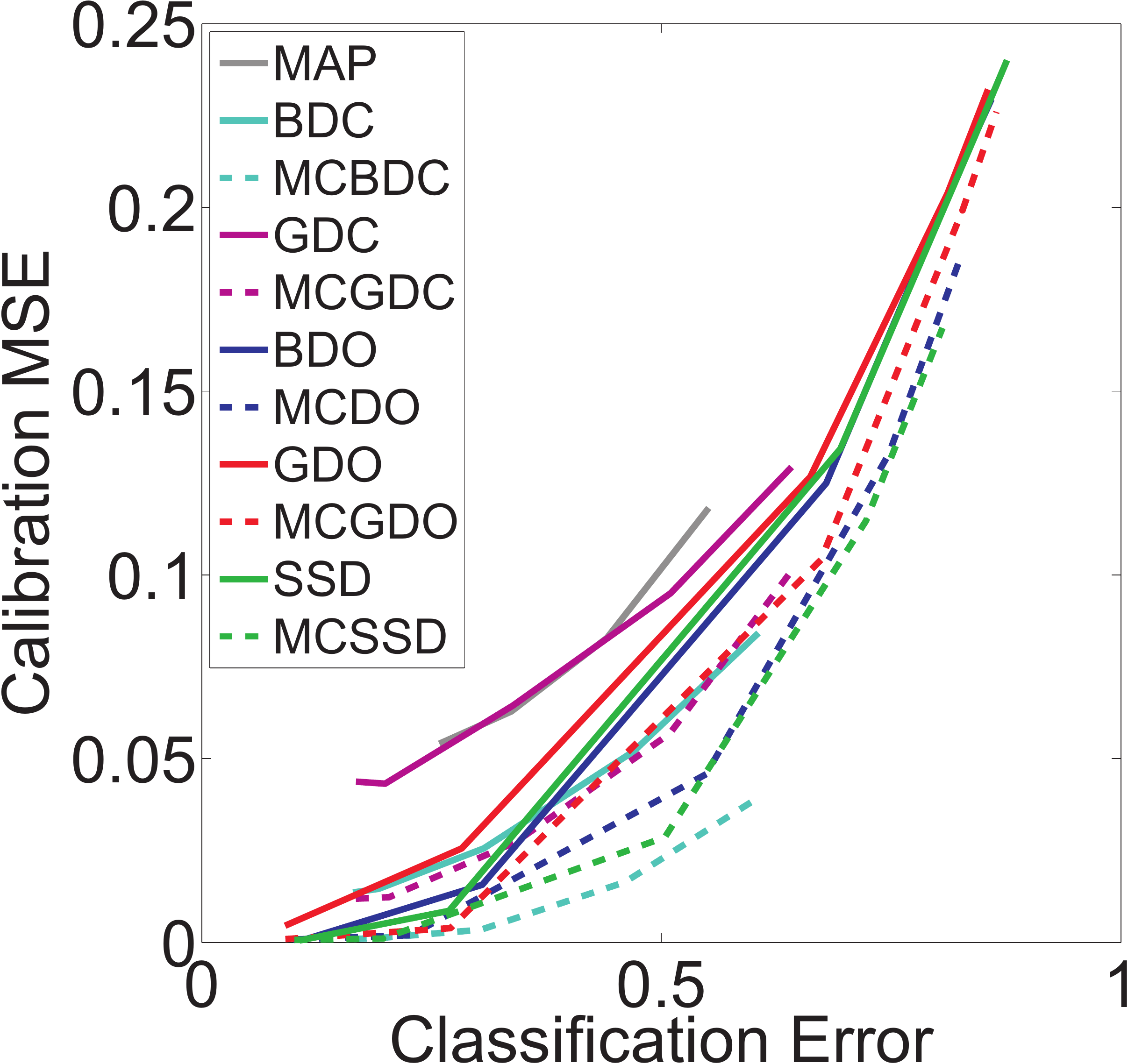}
        \caption{}
        \label{calibration_vs_error}
    \end{subfigure}
    \caption{The CIFAR-10 (a) classification error for additive Gaussian noise using standard deviations St.D of 0, 0.25, 0.5, 0.75, and 1, (b) mean squared error (MSE) between the $x = y$ line and the calibration plot (i.e. the frequency of the true label vs predicted probability of that label) for varying Gaussian image noise StD., and (c) calibration MSE versus the classification error for predicitons across all noise StD. for Bernoulli dropconnect (BDC), Gaussian dropconnect (GDC), Bernoulli dropout (BDO), Gaussian dropout (GDO), and spike-and-slab dropout (SSD) with and without MC sampling using 10 samples.}\label{plots}
\end{figure}

\begin{figure}
\begin{center}
\begin{subfigure}{\textwidth}
\includegraphics[width = \textwidth]{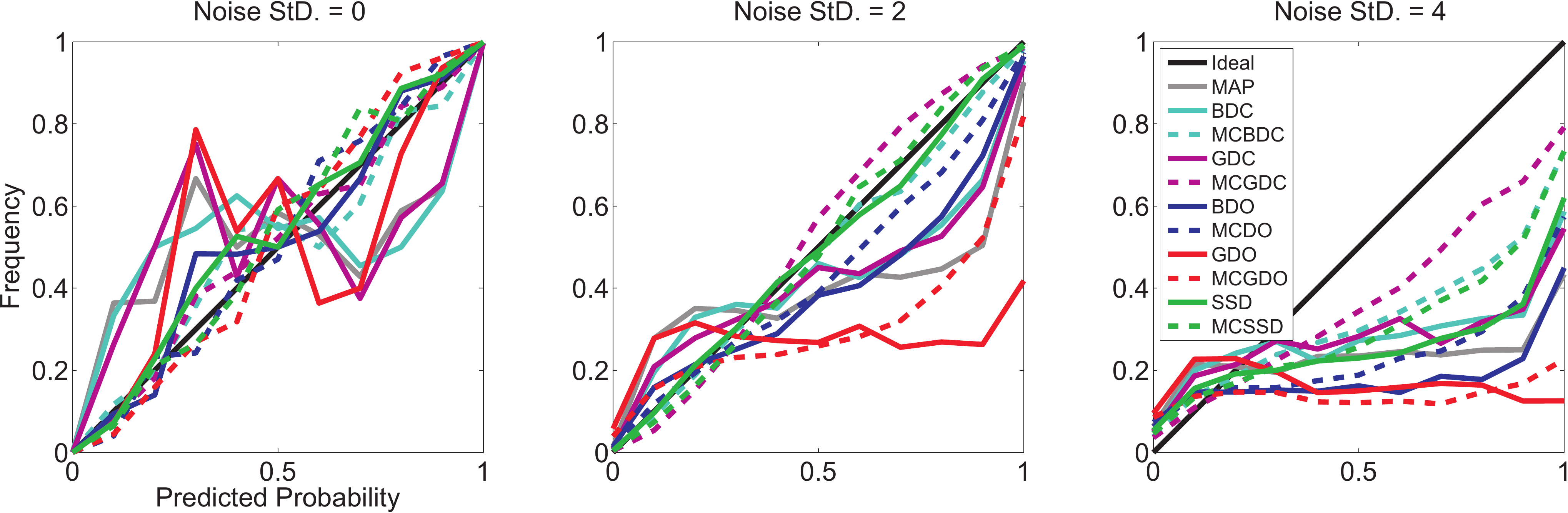}
\vskip -0.1in
        \caption{}
\end{subfigure}
\vskip 0.2in
\begin{subfigure}{\textwidth}
\includegraphics[width = \textwidth]{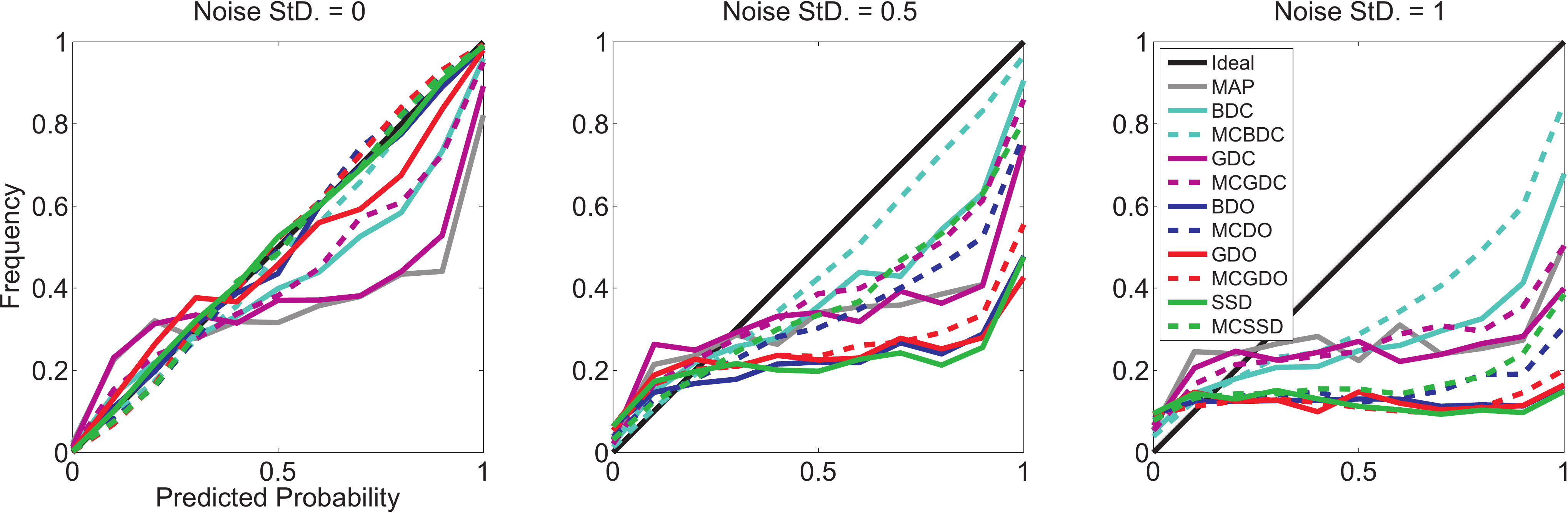}
\vskip -0.1in
        \caption{}
\end{subfigure}
\end{center}
    
    \caption{The $x = y$ line (Ideal) and the calibration plot (i.e. the frequency of the true label vs predicted probability of that label) for varying Gaussian image noise StD. for the (a) MNIST or (b) CIFAR-10 trained Bernoulli dropconnect (BDC), Gaussian dropconnect (GDC), Bernoulli dropout (BDO), Gaussian dropout (GDO), and spike-and-slab dropout (SSD) networks with and without MC sampling using 10 samples.}
\label{calibration_curves}
\end{figure}

While the dropout-based methods were the most accurate on the test-set (Table \ref{cnn-test-errors}), as image noise was added they became increasingly worse compared to the dropconnect-based networks (Figure \ref{mnist_plots}.a and \ref{plots}.a).  Sampling only consistently improved the accuracy for Bernoulli and spike-and-slab dropout. However, sampling did consistently improve the calibration of the networks as the image noise was increased (Figure \ref{mnist_plots}.b and \ref{plots}.b). For a given accuracy across each set of noisy test images, sampling also generally lead to better calibration (Figure \ref{mnist_plots}.c, \ref{plots}.c, and \ref{calibration_curves}). (See Supplementary Material for the calibration plots.) Gaussian dropout led to the highest test set accuracy, but it also led to reduced robustness to noise. While slightly less accurate on the test set, Bernoulli dropout and spike-and-slab dropout were much more robust.

Seemingly contradictory results have been reported in the literature regarding CIFAR-10 and MC Bernoulli dropout. \citet{gal2015dropout} found that standard Bernoulli dropout methods led to relatively inaccurate networks when dropout was used at every layer in a CNN, whereas MC sampling increased the accuracy of these networks. However, \citet{srivastava2014dropout} found that using dropout at every layer led to increased generalization performance even without sampling at prediction time. In our CIFAR-10 experiments, but not our MNIST experiments, we have found that using sampling at prediction time makes networks more robust to high variance dropout. Using lower variance dropout results in standard and MC methods having similar accuracies, while using higher variance distributions results in MC inference outperforming standard methods (Figure \ref{sampling_p_curves}). (See Supplementary Material for more results when using $p=0.5$.) These results indicate that Bernoulli or Gaussian dropout with MC sampling are less dependent on the exact value of $p$ and can allow higher levels of dropout regularization to be used.

\begin{figure}[b]
\centering
    \includegraphics[width=0.5\textwidth]{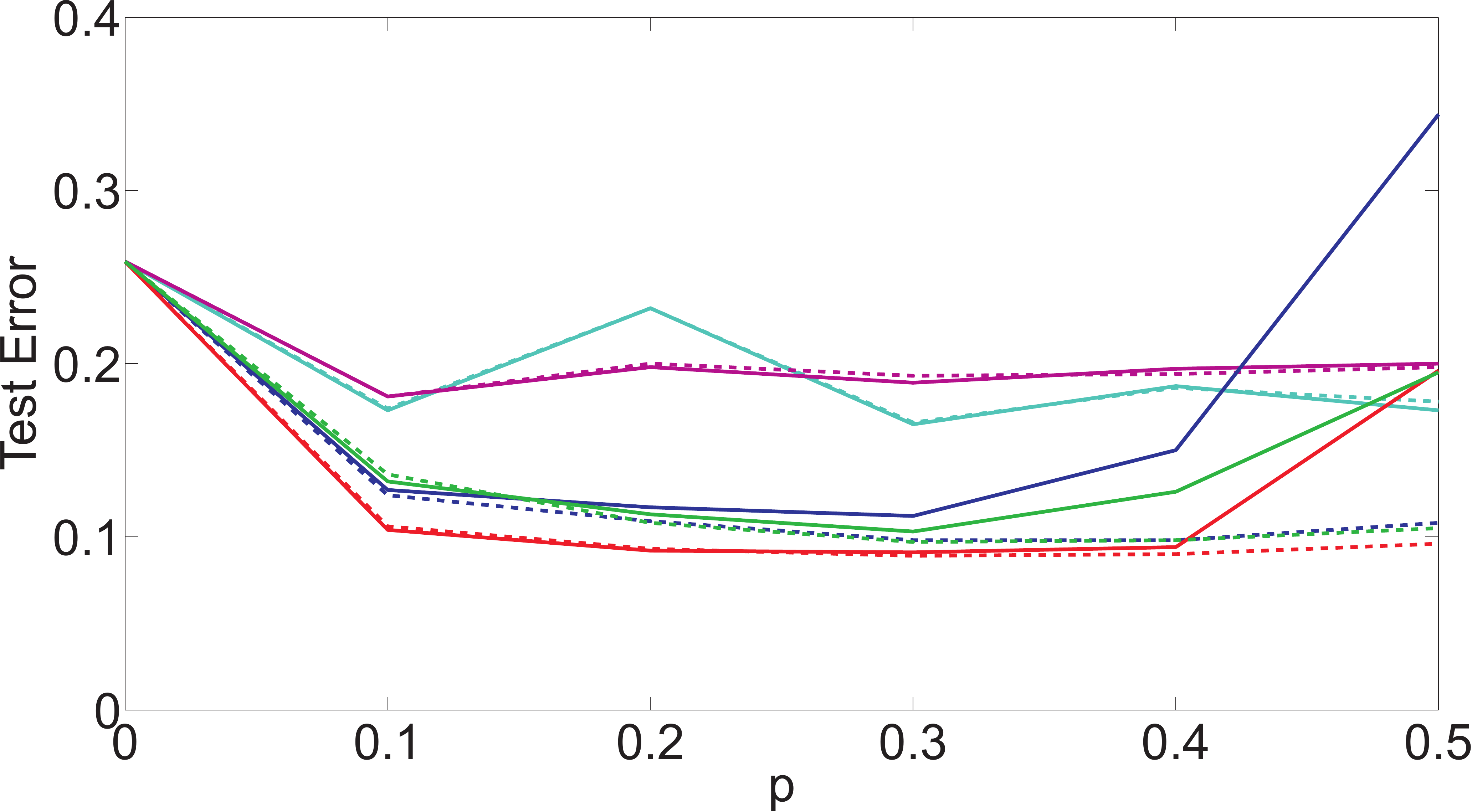}
    \caption{The classification error for the CIFAR-10 test set using different $p$ hyperparameter values. For spike-and-slab dropout, $p_{do}$ was varied, while $p_{dc}$ was fixed.}
    \label{sampling_p_curves}
\end{figure}

\section{Discussion}
\label{discussion}

L2 regularization and Bernoulli dropout are widely used for regularization and routinely lead to increased testing accuracy. However, the uncertainty learned do not generalize well. However, performing approximate Bayesian inference via sampling during training and testing allowed CNNs to better model their uncertainty. Dropconnect-based CNNs performed worse on the unmodified test set, but were much more robust to deviations from the training distribution. On the other hand, dropout-based networks, particularly MC Gaussian dropout, performed well on the unmodified test set, but were not as robust. Using sampling and combining Bernoulli dropout and Gaussian dropconnect to approximate the use of spike-and-slab variational distributions lead to a CNN that performed better near the test set than the dropconnect methods and more robustly represented its uncertainty compared to the dropout methods.

\section{Acknowledgments}
\label{acknowledgements}

The authors would like to thank Sergii Strelchuk, Charles Zheng, Yarin Gal, Richard Turner, and Aapo Hyv\"{a}rinen for their comments on previous versions of the manuscript. This research was funded by the UK Medical Research Council (Programme MC-A060- 5PR20), by a European Research Council Starting Grant (ERC-2010-StG 261352), and by the Human Brain Project (EU grant 604102 'Context-sensitive multisensory object recognition: a deep network model constrained by multi-level, multi-species data').

\bibliographystyle{plainnat}
\bibliography{nips_2017}

\begin{thebibliography}{26}
\providecommand{\natexlab}[1]{#1}
\providecommand{\url}[1]{\texttt{#1}}
\expandafter\ifx\csname urlstyle\endcsname\relax
  \providecommand{\doi}[1]{doi: #1}\else
  \providecommand{\doi}{doi: \begingroup \urlstyle{rm}\Url}\fi

\bibitem[Barber and Bishop(1998)]{barber1998ensemble}
David Barber and Christopher~M Bishop.
\newblock Ensemble learning in bayesian neural networks.
\newblock \emph{NATO ASI SERIES F COMPUTER AND SYSTEMS SCIENCES}, 168:\penalty0
  215--238, 1998.

\bibitem[Blundell et~al.(2015)Blundell, Cornebise, Kavukcuoglu, and
  Wierstra]{blundell2015weight}
Charles Blundell, Julien Cornebise, Koray Kavukcuoglu, and Daan Wierstra.
\newblock Weight uncertainty in neural network.
\newblock In \emph{Proceedings of The 32nd International Conference on Machine
  Learning}, pages 1613--1622, 2015.

\bibitem[Gal(2016)]{Gal2016Uncertainty}
Yarin Gal.
\newblock \emph{Uncertainty in Deep Learning}.
\newblock PhD thesis, University of Cambridge, 2016.

\bibitem[Gal and Ghahramani(2015)]{gal2015dropout}
Yarin Gal and Zoubin Ghahramani.
\newblock Dropout as a bayesian approximation: Insights and applications.
\newblock In \emph{Deep Learning Workshop, ICML}, 2015.

\bibitem[Gal and Ghahramani(2016)]{gal2016bayesian}
Yarin Gal and Zoubin Ghahramani.
\newblock Bayesian convolutional neural networks with {B}ernoulli approximate
  variational inference.
\newblock In \emph{4th International Conference on Learning Representations
  (ICLR) workshop track}, 2016.

\bibitem[George and McCulloch(1997)]{george1997approaches}
Edward~I George and Robert~E McCulloch.
\newblock Approaches for bayesian variable selection.
\newblock \emph{Statistica sinica}, pages 339--373, 1997.

\bibitem[Graves(2011)]{graves2011practical}
Alex Graves.
\newblock Practical variational inference for neural networks.
\newblock In \emph{Advances in Neural Information Processing Systems}, pages
  2348--2356, 2011.

\bibitem[Hern{\'a}ndez-Lobato and Adams(2015)]{hernandez2015probabilistic}
Jos{\'e}~Miguel Hern{\'a}ndez-Lobato and Ryan Adams.
\newblock Probabilistic backpropagation for scalable learning of bayesian
  neural networks.
\newblock In \emph{International Conference on Machine Learning}, pages
  1861--1869, 2015.

\bibitem[Hinton and Van~Camp(1993)]{hinton1993keeping}
Geoffrey~E Hinton and Drew Van~Camp.
\newblock Keeping the neural networks simple by minimizing the description
  length of the weights.
\newblock In \emph{Proceedings of the sixth annual conference on Computational
  learning theory}, pages 5--13. ACM, 1993.

\bibitem[Ishwaran and Rao(2005)]{ishwaran2005spike}
Hemant Ishwaran and J~Sunil Rao.
\newblock Spike and slab variable selection: frequentist and bayesian
  strategies.
\newblock \emph{Annals of Statistics}, pages 730--773, 2005.

\bibitem[Jyl{\"a}nki et~al.(2014)Jyl{\"a}nki, Nummenmaa, and
  Vehtari]{jylanki2014expectation}
Pasi Jyl{\"a}nki, Aapo Nummenmaa, and Aki Vehtari.
\newblock Expectation propagation for neural networks with sparsity-promoting
  priors.
\newblock \emph{Journal of Machine Learning Research}, 15\penalty0
  (1):\penalty0 1849--1901, 2014.

\bibitem[Kendall and Gal(2017)]{kendall2017uncertainties}
Alex Kendall and Yarin Gal.
\newblock What uncertainties do we need in bayesian deep learning for computer
  vision?
\newblock \emph{arXiv preprint arXiv:1703.04977}, 2017.

\bibitem[Kingma et~al.(2015)Kingma, Salimans, and
  Welling]{kingma2015variational}
Diederik~P Kingma, Tim Salimans, and Max Welling.
\newblock Variational dropout and the local reparameterization trick.
\newblock In \emph{Advances in Neural Information Processing Systems}, pages
  2575--2583, 2015.

\bibitem[Krizhevsky and Hinton(2009)]{krizhevsky2009learning}
Alex Krizhevsky and Geoffrey Hinton.
\newblock Learning multiple layers of features from tiny images, 2009.

\bibitem[Krizhevsky et~al.(2012)Krizhevsky, Sutskever, Hinton, Alex, Sutskever,
  and Hinton]{Krizhevsky2012}
Alex Krizhevsky, Ilya Sutskever, Geoffrey~E Hinton, Krizhevsky Alex, Ilya
  Sutskever, and Geoffrey~E Hinton.
\newblock {ImageNet Classification with Deep Convolutional Neural Networks}.
\newblock In \emph{Advances In Neural Information Processing Systems}, pages
  1--9, 2012.
\newblock ISBN 9781627480031.
\newblock URL
  \url{http://papers.nips.cc/paper/4824-imagenet-classification-with-deep-convolutional-neural-networks.pdf}.

\bibitem[LeCun et~al.(1998)LeCun, Bottou, Bengio, and
  Haffner]{lecun1998gradient}
Yann LeCun, L{\'e}on Bottou, Yoshua Bengio, and Patrick Haffner.
\newblock Gradient-based learning applied to document recognition.
\newblock \emph{Proceedings of the IEEE}, 86\penalty0 (11):\penalty0
  2278--2324, 1998.

\bibitem[Louizos(2015)]{louizos2015smart}
Christos Louizos.
\newblock Smart regularization of deep architectures.
\newblock Master's thesis, University of Amsterdam, 2015.

\bibitem[MacKay(1992)]{mackay1992practical}
David~JC MacKay.
\newblock A practical bayesian framework for backpropagation networks.
\newblock \emph{Neural computation}, 4\penalty0 (3):\penalty0 448--472, 1992.

\bibitem[Madigan and Raftery(1994)]{madigan1994model}
David Madigan and Adrian~E Raftery.
\newblock Model selection and accounting for model uncertainty in graphical
  models using occam's window.
\newblock \emph{Journal of the American Statistical Association}, 89\penalty0
  (428):\penalty0 1535--1546, 1994.

\bibitem[Mitchell and Beauchamp(1988)]{mitchell1988bayesian}
Toby~J Mitchell and John~J Beauchamp.
\newblock Bayesian variable selection in linear regression.
\newblock \emph{Journal of the American Statistical Association}, 83\penalty0
  (404):\penalty0 1023--1032, 1988.

\bibitem[Mnih et~al.(2015)Mnih, Kavukcuoglu, Silver, Rusu, Veness, Bellemare,
  Graves, Riedmiller, Fidjeland, Ostrovski, Petersen, Beattie, Sadik,
  Antonoglou, King, Kumaran, Wierstra, Legg, and Hassabis]{Mnih2015}
Volodymyr Mnih, Koray Kavukcuoglu, David Silver, Andrei~a Rusu, Joel Veness,
  Marc~G Bellemare, Alex Graves, Martin Riedmiller, Andreas~K Fidjeland, Georg
  Ostrovski, Stig Petersen, Charles Beattie, Amir Sadik, Ioannis Antonoglou,
  Helen King, Dharshan Kumaran, Daan Wierstra, Shane Legg, and Demis Hassabis.
\newblock {Human-level control through deep reinforcement learning}.
\newblock \emph{Nature}, 518\penalty0 (7540):\penalty0 529--533, 2015.
\newblock ISSN 0028-0836.
\newblock \doi{10.1038/nature14236}.
\newblock URL \url{http://dx.doi.org/10.1038/nature14236}.

\bibitem[Neal(2012)]{neal2012bayesian}
Radford~M Neal.
\newblock \emph{Bayesian learning for neural networks}, volume 118.
\newblock Springer Science \& Business Media, 2012.

\bibitem[Silver et~al.(2016)Silver, Huang, Maddison, Guez, Sifre, Van
  Den~Driessche, Schrittwieser, Antonoglou, Panneershelvam, Lanctot,
  et~al.]{silver2016mastering}
David Silver, Aja Huang, Chris~J Maddison, Arthur Guez, Laurent Sifre, George
  Van Den~Driessche, Julian Schrittwieser, Ioannis Antonoglou, Veda
  Panneershelvam, Marc Lanctot, et~al.
\newblock Mastering the game of go with deep neural networks and tree search.
\newblock \emph{Nature}, 529\penalty0 (7587):\penalty0 484--489, 2016.

\bibitem[Srivastava et~al.(2014)Srivastava, Hinton, Krizhevsky, Sutskever, and
  Salakhutdinov]{srivastava2014dropout}
Nitish Srivastava, Geoffrey Hinton, Alex Krizhevsky, Ilya Sutskever, and Ruslan
  Salakhutdinov.
\newblock Dropout: A simple way to prevent neural networks from overfitting.
\newblock \emph{The Journal of Machine Learning Research}, 15\penalty0
  (1):\penalty0 1929--1958, 2014.

\bibitem[Wan et~al.(2013)Wan, Zeiler, Zhang, Cun, and
  Fergus]{wan2013regularization}
Li~Wan, Matthew Zeiler, Sixin Zhang, Yann~L Cun, and Rob Fergus.
\newblock Regularization of neural networks using dropconnect.
\newblock In \emph{Proceedings of the 30th International Conference on Machine
  Learning (ICML-13)}, pages 1058--1066, 2013.

\bibitem[Welling and Teh(2011)]{welling2011bayesian}
Max Welling and Yee~W Teh.
\newblock Bayesian learning via stochastic gradient langevin dynamics.
\newblock In \emph{Proceedings of the 28th International Conference on Machine
  Learning (ICML-11)}, pages 681--688, 2011.

\end{thebibliography}

\includepdf[pages={-}]{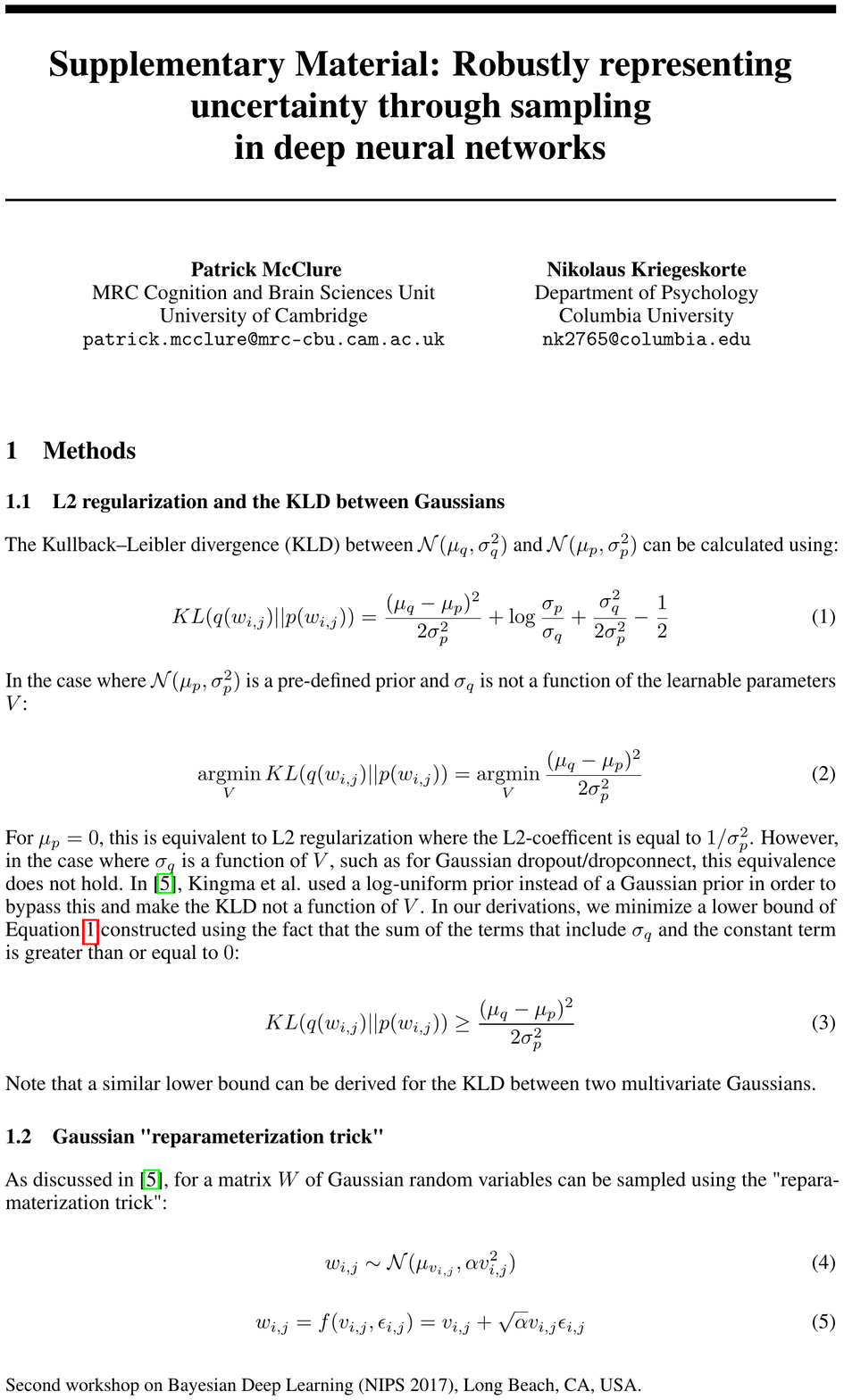}

\end{document}